\documentclass[lettersize,journal]{IEEEtran}    

\IEEEoverridecommandlockouts                              

\usepackage{graphicx} 
\usepackage{subfigure}  
\usepackage{overpic}  
\usepackage{float} %
\usepackage{stfloats}

\usepackage{tabu} 
\usepackage{multirow}
\usepackage{makecell}    
\usepackage{array}

\usepackage{amsfonts}
\usepackage{amsmath, amssymb, mathtools}
\usepackage{algorithm}
\usepackage{algpseudocode}
\usepackage{dsfont}
\usepackage[noadjust]{cite}
\usepackage[normalem]{ulem}
\usepackage[colorlinks,linkcolor=red]{hyperref}
\usepackage{balance}
\usepackage{xcolor}

\newcommand{\ie}{\emph{i.e.}}

\newcommand{\sr}[1]{\scriptsize{\textcolor{purple}{#1}}}
\newcommand{\sg}[1]{\scriptsize{\textcolor{teal}{#1}}}
\newcommand\mypara[1]{\vspace{0mm}\noindent\textbf{#1}}

\title{\LARGE \bf
Efficient 3D Perception on Multi-Sweep Point Cloud \\ with Gumbel Spatial Pruning
}

\author{Tianyu Sun$^{\ast}$, Jianhao Li$^{\ast}$, Xueqian Zhang, Zhongdao Wang, Bailan Feng, Hengshuang Zhao$^{\dagger}$ 
\thanks{Tianyu Sun and Xueqian Zhang are with the Department of Electronic Engineering, Tsinghua University, China.}
\thanks{Jianhao Li is with the Department of Computer Science and Engineering, Beihang University, China.}
\thanks{Zhongdao Wang and Bailan Feng are with Noah's Ark Lab, Beijing, China.}
\thanks{Hengshuang Zhao is with the Department of Computer Science, University of Hong Kong, China.}
\thanks{* Equally Contributed}
\thanks{$\dagger$ Corresponding Author}
}

\begin{document}

\maketitle
\thispagestyle{empty}
\pagestyle{empty}

\begin{abstract}
This paper studies point cloud perception within outdoor environments. Existing methods face limitations in recognizing objects located at a distance or occluded, due to the sparse nature of outdoor point clouds.
In this work, we observe a significant mitigation of this problem by accumulating multiple temporally consecutive point cloud sweeps, resulting in a remarkable improvement in perception accuracy. However, the computation cost also increases, hindering previous approaches from utilizing a large number of point cloud sweeps.
To tackle this challenge, we find that a considerable portion of points in the accumulated point cloud is redundant, and discarding these points has minimal impact on perception accuracy. We introduce a simple yet effective Gumbel Spatial Pruning (GSP) layer that dynamically prunes points based on a learned end-to-end sampling. The GSP layer is decoupled from other network components and thus can be seamlessly integrated into existing point cloud network architectures.
Without incurring additional computational overhead, we increase the number of point cloud sweeps from 10, a common practice, to as many as 40. Consequently, there is a significant enhancement in perception performance. For instance, in nuScenes 3D object detection and BEV map segmentation tasks, our pruning strategy improves several 3D perception baseline methods.
\end{abstract}

\section{INTRODUCTION}

\begin{figure}
	\centering
	\includegraphics[width=0.95\linewidth]{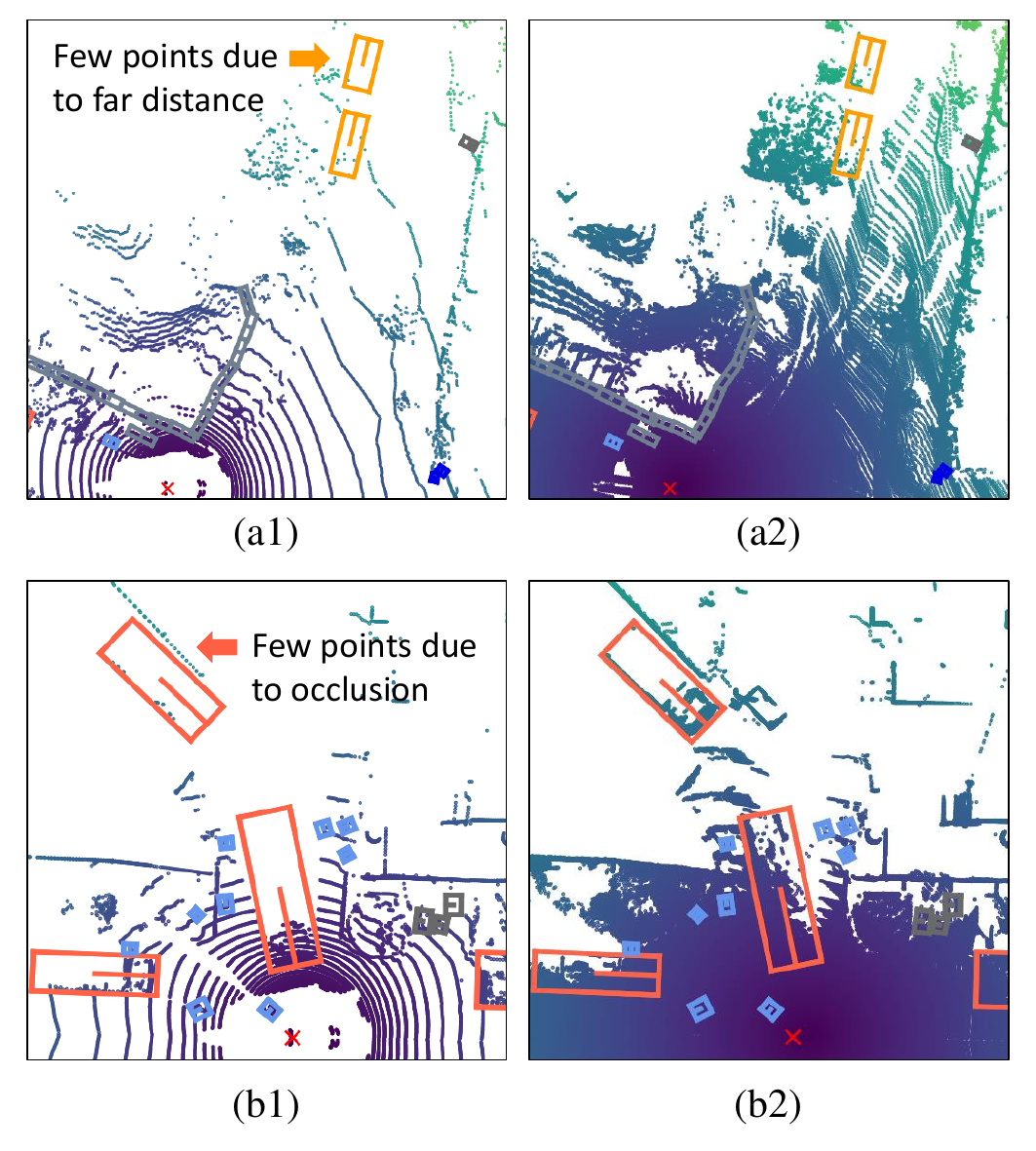}
	\caption{
		Sparse or missing outdoor point cloud data poses significant challenges for perception tasks, particularly at a distance (a1) or when objects are occluded (b1). Accumulating multiple temporally consecutive point cloud sweeps enhances the density of the point cloud (a2) (b2), thereby facilitating perception. However, this improvement comes at the expense of increased computational cost. This work explores cost-effective strategies for leveraging multiple point cloud sweeps.
	}
	\label{fig:intro}
\end{figure}

The past several years have witnessed exponential growth in 3D perception techniques~\cite{2017VoxelNet,2019PointRCNN,jin2025angledomainguidancelatent,sun2024diffusion}, including considerable point cloud-based methods~\cite{2020Point,Yi_2023_ICCV}, primarily attributed to the rapid advancement and significant cost reduction of point cloud sensors.  In outdoor applications, such as autonomous driving~\cite{li2025segment} and robotics~\cite{hu2025variation,xie2023part}, the utilization of point clouds is essential for precise scene geometry measurement and dynamic object localization. Typically, a point cloud sensor is positioned atop the ego-car and performs a comprehensive~\cite{Caesar_2020_CVPR} or partial~\cite{Sun_2020_CVPR} panoramic scan of the surrounding environment.
However, this operational mechanism presents two inherent limitations. Firstly, the density of captured points exhibits a gradient, with higher density near the ego-car and sparser distribution at greater distances, thereby posing challenges for distant perception (Fig.~\ref{fig:intro} a1). Secondly, occluded objects remain undetected by the point cloud sensor due to their limited or absent beam reflections (Fig.~\ref{fig:intro} b1).
Consequently, these limitations significantly hinder the accuracy and reliability of point-cloud-based perception. To overcome these challenges, previous endeavors have introduced vision cues and pursued the fusion of data from multiple sensor modalities~\cite{Li_2022_CVPR,sun2023trosd, liu2022bevfusion}.
In a broader context, leveraging multi-view information has proven highly effective in resolving spatial occlusions and enhancing geometric consistency, benefiting various visual tasks~\cite{sun2026mvanimate,yi2024mvgamba}.

Nevertheless, even when considering point cloud data as the sole input, there exist approaches to mitigate the aforementioned issues.
Autonomous vehicles are typically equipped with Inertial Measurement Units (IMUs) that can measure the pose of the ego-car. Leveraging this information, we can align and accumulate multiple temporally adjacent point cloud sweeps into the current frame (Figure~\ref{fig:intro} a2 \& b2). As a result, the sparse regions of the point cloud become denser, thereby facilitating the subsequent perception step.
Despite the lack of a comprehensive investigation, this technique has been widely adopted as a common practice in point cloud-based tasks, such as object detection~\cite{Jiao_2023_CVPR,Li_2022_CVPR,transfusion,chen2023largekernel3d} and semantic segmentation~\cite{10204261,2020Sparse,Zhou_2021_CVPR,jiang2021guided} due to its consistent advantages.

The flip side of the coin is the additional time consumption and computation cost that grow linearly with the rising number of sweeps. To strike a balance between network efficiency and performance gains, a typical trade-off is made by setting the number of sweeps to 10 in the nuScenes dataset~\cite{Caesar_2020_CVPR,qi2023ocbev,wang2023uni3detr,lu2023link,ge2023metabev,liu2022prototype} and 4 in the Waymo Open datase~\cite{Sun_2020_CVPR,guan2022m3detr,fan2022fully,Liu_2023_ICCV,ye2022lidarmutlinet,li2023frame}. 
Notably, our observations reveal that augmenting the number of sweeps beyond the common practice further enhances perception performance. However,  the increasing computation cost hinders the exploration of this possibility.

In this work, our objective is to enhance point cloud perception by leveraging a greater number of temporal sweeps while minimizing any additional computational overhead. To achieve this, we make use of an observation that the accumulated point cloud data exhibits high redundancy ( Fig.~\ref{fig:intro}). For instance, the road surface in close proximity to the ego-car often contains a significant number of densely packed points. Intuitively, if we can identify and selectively discard such regions, computational costs can be dramatically reduced without significantly impacting the performance of the tasks at hand.

In light of the above consideration, we propose a simple yet efficient Gumbel Spatial Pruning (GSP) Layer to fulfill the dynamic pruning of point clouds.
The GSP layer is a plug-and-play component that can be readily inserted after each computation layer. It learns a binary classifier through Gumbel Softmax to decide which points should be pruned. Our key finding is that it is crucial to use \emph{hard} Gumbel Softmax which outputs discrete binary values rather than continuous logits. This way, we can use the discrete prediction for indexing and avoid thresholding the probability map during inference, keeping the data distribution consistent with training, and thus minimizing the performance loss brought by pruning. GSP layer uses neither additional supervision like pre-computed pruning label, nor prior cues such as feature magnitude~\cite{liu2022spatial}.  The pruning is learned solely from the supervision of the task loss and a sparse regularization, in an end-to-end manner. Extensive experiments show that, by introducing the GSP layer, the sweep number can be enlarged to $4\times$ without incurring additional computational overhead, either in terms of FLOPS or latency, and perception accuracy remarkably improves. In the nuScenes dataset, we improve several 3D perception networks, including the vanilla TransL \cite{transfusion} and LargeKernel3D \cite{chen2023largekernel3d}, on both 3D object detection tasks and BEV map segmentation tasks. Compared with existing spatial pruning methods~\cite{liu2022spatial}, we achieve a better trade-off between task performance and network efficiency. 

To summarize, our contribution is three-fold.
\begin{itemize}
	\item  We present that more temporally accumulated sweeps beyond the common practice still bring considerable gains in perception accuracy. 
	\item We propose a novel Gumbel Spatial Pruning Layer, which reduces the computational cost by $4\times$ with a neglectable performance drop. 
	\item Equipped with the GSP layer, we train 3D detection and segmentation models with $4\times$ sweeps with no additional computation cost. The accuracy for each task significantly improves, even on the basis of strong SOTA models.
\end{itemize}


\section{Related Work}


The point cloud is often utilized in the area of 3D perception tasks, carrying a great amount of spatial information. Recent researchers have focused on retrieving effective information from limited point cloud samples~\cite{yi2024mvgamba,liang2024pointmamba}. In this section, we will introduce the up-to-date work on point cloud perception and point cloud pruning. 

\mypara{Single-sweep point cloud perception.}
At the early stage of point cloud implementation, its usage often requires matching with RGB images to leverage the missing spatial information of a 2D input. 
In this case, a single sweep is merely an assistive input. Similarly, single sweep perception is also implemented in point-cloud-only perception tasks. 
Yan et al.~\cite{2020Sparse} propose JS3C-Net for single sweep point cloud segmentation via learning contextual shape priors. Barrera et al.~\cite{2021barrera} introduce BirdNet+ to deal with object detection with a single point cloud sweep. 
However, these researches all fail to densify the target point cloud and have been gradually surpassed by multi-sweep methods in 3D perception tasks. 
Especially in outdoor scenes, a single sweep can hardly provide enough points to fully perceive the environment. 
For relatively small-scale target objects, the sparse point cloud would cause distortion or even absence. 
This weakness brings out the utility of multi-sweep perception. 

\mypara{Multi-sweep point cloud perception.}
Multi-sweep perception was proposed because of the overwhelming point cloud information, especially in autonomous driving areas, where the combination of vehicle and LiDAR makes point cloud collection much easier. For datasets such as nuScenes~\cite{Caesar_2020_CVPR} and Waymo~\cite{Sun_2020_CVPR}, researchers tend to include multi-sweep point clouds in their perception modules and carry out various tasks. Lu et al.~\cite{lu2023link} propose a LinK-based backbone to solve detection and segmentation tasks. 
Multi-sweep perception effectively promotes the performance of various algorithms, but the increasing workload due to massive input leads to larger time and computing resource consumption. Liu et al.~\cite{Liu_2023_ICCV} propose MV-DeepSDF to reconstruct 3D vehicles from multi-sweep point clouds. 
Wang et al.~\cite{wang_2022_cvpr} construct PointMotionNet which learns from a sequence of 3D point clouds and performs well in multi-sweep semantic segmentation tasks. 
However, almost all mentioned measures are faced with increasing time and computing power consumption, compared with single-sweep algorithms. 

\mypara{Point cloud pruning}
Point cloud pruning is a pre-processing stage of the 3D perception model. During a pruning process, unnecessary point clouds should be eliminated and the rest should be kept. The goal of pruning strategies is generally keeping the performance at a high level while lowering the time and computation consumption. 

Most pruning methods focus on the FLOPs or the latency of the network because these two metrics can directly show the time consumption or the number of operations. Liu et al.~\cite{liu2022spatial} design Spatial Pruned Sparse Convolution(SPS-Conv) for efficient 3D detection tasks, achieving a reduction in FLOPs without compromising performance. Huang et al.~\cite{huang2023cp3} propose {CP}$^{3}$, a channel-pruning method for 3D point-based neural networks, with a drop in FLOPs in several 3D tasks. Zhou et al.~\cite{Zhou_2021_CVPR} introduce self-adversarial pruning(SAP), a point cloud pruning method for 3D detection tasks, achieving great progress in lowering the inference latency of the model. Tang et al.~\cite{tang2022torchsparse} propose Torchsparse, an efficient point cloud engine, which only optimizes inference latency and trades FLOPs for computation regularity. 

However, most of the mentioned pruning strategies fail to lower both latency and FLOPs with the main metrics remaining at the same level. Whereas, both of these metrics are essential for building an efficient perception network. Therefore, this target is essential to defining a successful pruning method. 


\section{Background and Motivation}
Due to the unbounded nature of outdoor scenes, The point cloud obtained by a roof-mounted point cloud sensor in autonomous vehicles,  exhibits a gradient distribution, with higher point density in proximity to the ego-car and a sparser distribution at greater distances, thereby presenting challenges for distant perception. 
Besides, the point cloud's laser beams capture only the first intersecting surface along their direction, rendering occluded objects unperceivable. In prior arts~\cite{liu2022bevfusion,transfusion,chen2023voxelnext,2020Sparse,wang2023uni3detr}, a common practice to mitigate this issue is to accumulate a history point cloud sweeps into the current point cloud. Given the ego-poses at the current timestamp $t$ and a history timestamp $t - \Delta t$, together with the extrinsic parameters of the point cloud sensor, we obtain the transformation matrix $P_{t-\Delta t, t}$ and then transform the history point cloud by
        $(\mathbf{x'}_{t-\Delta t},
        1)^{T} 
    = P_{t-\Delta t, t} 
        (\mathbf{x}_{t-\Delta t},
        1)^{T} $, where $\mathbf{x}_{t-\Delta t}$ indicates the original spatial coordinate of a point and $\mathbf{x'}_{t-\Delta t}$ is its aligned coordinate in the current timestamp.
Multiple sweeps are then concatenated, forming an accumulated point cloud $\{\mathbf{x'}_{t-\Delta t} \in \mathbf{X'}_{t-\Delta t} \vert \Delta t =0,1,2,...,T\}$. To specify which timestep each point comes from, an additional temporal dimension is appended after the spatial coordinates. Consequently, we have 4D coordinates represented by $(\mathbf{x'}_{t-\Delta t}, \Delta t)^T$.

A typical choice of the sweep number is a mild $T=10$ as used in most previous point-cloud-based object detectors~\cite{centerpoint,transfusion,liu2022bevfusion}, which means a duration of 0.4 seconds for 25FPS streaming data, and the accumulated point cloud is of size around $10^5$. 
Intuitively, further enlarging $T$ may still bring additional improvements in perception performance, but also results in a dramatic increase in computation cost, and thus is well explored in existing works. In this paper, we validate the existence of these benefits by imposing more temporal sweeps beyond the common setting. More importantly, we introduce a new method that significantly reduces computation and enables us to enjoy such benefits at almost no cost.



\begin{figure}[t]
\centering
\includegraphics[width=\linewidth]{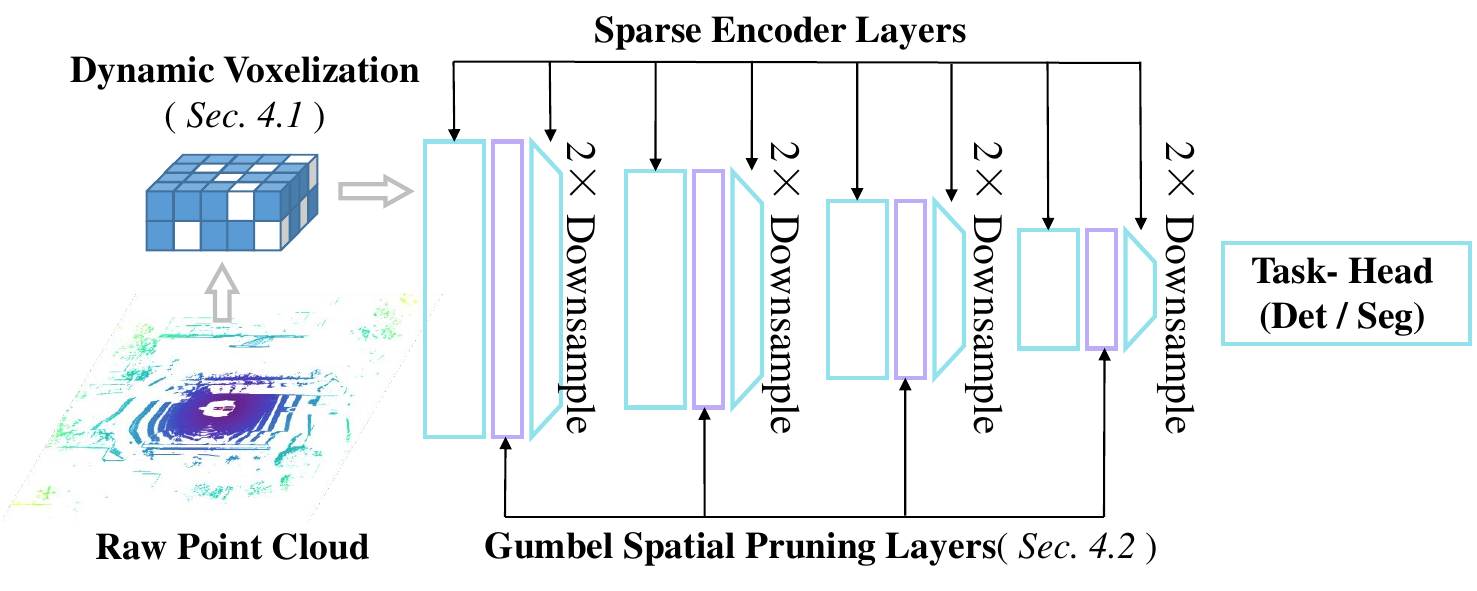}
\caption{\textbf{Overview of our pruning strategy.} The input point cloud is first processed by dynamic voxelization and then goes through the sparse encoder layers. Spatial pruning takes place before each down-sampling layer. }
\label{fig:method}
\end{figure}

\begin{figure*}[t]
\centering
\includegraphics[width=\textwidth]{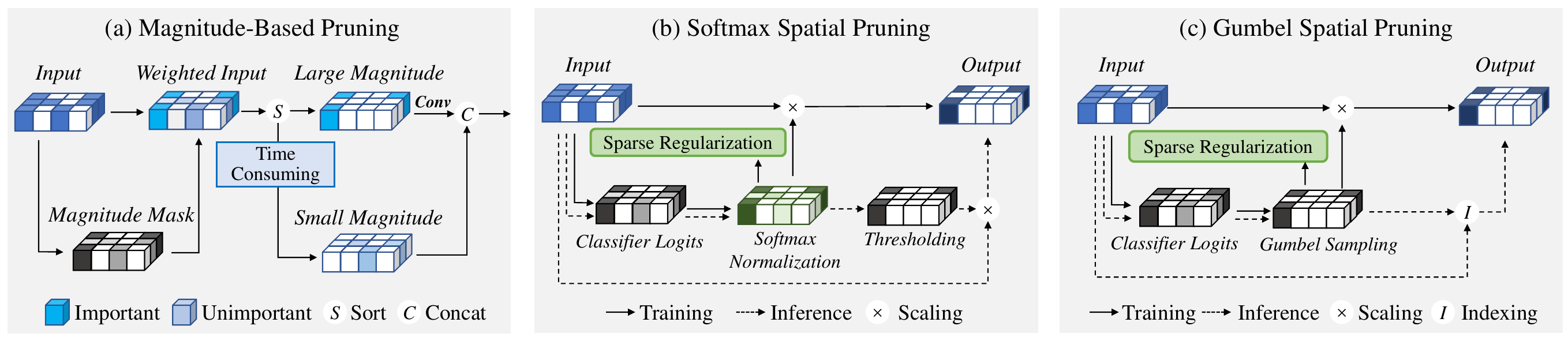}
\caption{\textbf{Comparison of different pruning methods.} (a) Magnitude-based pruning skips computation of low-magnitude features, but the sort operation it relies on brings additional cost. (b) Softmax spatial pruning learns to prune with sparse regularization, but the discrepancy between training and inference degenerates accuracy. (c) The proposed Gumbel spatial pruning is consistent between training and inference, boosting efficiency with almost no performance drop.}
\label{fig:pruning_comp}
\end{figure*}


\section{Approach}

Given the large redundancy of multi-sweep accumulated point clouds, we propose to prune fewer informative points to reduce computation. An overview of the proposed method is shown in Fig.~\ref{fig:method}. Following VoxelNeXt~\cite{chen2023voxelnext}, the spatial pruning takes place at the end of each network stage, before the downsampling layers. Furthermore, we suggest replacing the commonly used hard voxelization with a more efficient dynamic voxelization~\cite{zhou2019endtoend}, as the computational expense of the former becomes prohibitively high when dealing with an increased number of points resulting from the accumulation of multi-sweeps.

\subsection{Dynamic Voxelization}
In comparison to a single sweep of point cloud, the accumulated point cloud exhibits significantly higher redundancy, especially in the proximity of the ego-car. To mitigate this redundancy, voxelization is commonly employed following established practices~\cite{2017VoxelNet,second,centerpoint,transfusion}. Initially, the 3D space is partitioned into equidistant voxels, and the input points are grouped according to their voxel assignments. Subsequently, voxel features are encoded based on the points contained within each voxel. Existing methods typically adopt a \emph{hard voxelization}~\cite{2017VoxelNet} approach, which we have identified as having two limitations. Firstly, it imposes a predefined maximum point threshold per voxel, resulting in non-negligible information loss when excess points are rejected. Secondly, the implementation itself is inefficient, making it impractical to handle the increasing time consumption and computational cost when we accumulate more sweeps.

To address this issue, we find a simple \emph{dynamic voxelization}~\cite{zhou2019endtoend} strategy not only preserves more information of input points but also remarkably speeds up. Specifically, we record \emph{all} points inside a voxel instead of randomly sampling a pre-defined number as done in hard voxelization. Then, the voxel feature is simply encoded by averaging along the spatial dimensions and max-pooling along the temporal dimension, \textit{i.e.}, the voxel feature is $(\frac{1}{N}{\sum_{i=1}^N\mathbf{x}_i}, \max_{i=1}^{N}{\Delta t_i})^T$.
In practice, we alter to small voxel sizes in order to reduce information loss from averaging and max-pooling. 
An empirical comparison between hard \textit{v.s.} dynamic voxelization is shown in Section~\ref{ablation}.

\subsection{Spatially Voxel Pruning}
\label{sec:spatial_pruning}

After voxelization, we employ a voxel-based network to extract 3D features of the scene. Considering that the voxels are highly redundant in our accumulated point cloud, we explore the possibility of spatially pruning in the voxel space. In the following section, we first discuss the limitations of existing pruning strategies and then present our insights and improvements. A comparison is visualized in Fig.~\ref{fig:pruning_comp}

\paragraph{Magnitude-based Pruning.} 
A prevalent implementation of spatially voxel pruning resorts to hand-crafted pruning rules. Given a set of voxel features $\{f_i \}_{i=1}^{N}$, SPS-Conv~\cite{liu2022spatial} and VoxelNeXt~\cite{chen2023voxelnext} sort the features based on their magnitudes $ \vert f_i \vert$, and subsequently divide the set into two subsets. The first subset consists of the top half of features with larger magnitudes, which are then forwarded to subsequent layers for further computation. The other subset contains the remaining features and remains unprocessed until it is concatenated with the processed top half.

This pruning strategy demonstrates notable efficacy, enabling a considerable reduction of approximately $50\%$ in FLOPs while maintaining performance integrity. However, it has been observed in~\cite{chen2023voxelnext} that the reduction in actual inference speed is not as pronounced as the decrease in FLOPs. This discrepancy can be attributed to the fact that the computational cost of the sorting operation utilized during pruning is not accounted for in FLOPs calculation, while its impact on latency is non-negligible.
Additionally, despite its effectiveness,  magnitude-based pruning is apparently suboptimal as the rule is hand-crafted.

\paragraph{Softmax Spatial Pruning.} 
To tackle the above issues,
we present a plausible design that \emph{learns} a spatial pruning operation in an end-to-end manner.
In the pruning layer, a voxel feature $f_i$ undergoes a two-way classifier $\sigma(\cdot)$ which outputs two-dimensional logits. The logits are then Softmax-normalized. Specifically, we obtain $p ( f_i) = \frac{e^{\sigma_1 (f_i)} } {e^{\sigma_0 (f_i)} + e^{\sigma_1 (f_i)}} \in (0,1)$, where $\sigma_i(\cdot)$ indicates the $i$-th dimension of the logits, and $p(f_i)$ indicates the predicted probability of keeping this voxel. 
Subsequently, the voxel feature is scaled by the factor $p(f_i)$ and sent to the following layers. During training, this operation does not save computation. During inference, we threshold $p(f_i)$ by a value such as $0.5$ and then discard voxels with low probability, therefore achieving computation reduction. Note that no additional operation such as sorting is involved, making it possible for us to achieve consistent reduction on both inference latency and FLOPs.

In practice, we do not impose explicit supervision on the binary classifier, so a shortcut solution may emerge that the classifier predicts $p(f_i) \rightarrow 1$ for every $f_i$. Therefore, we introduce a sparse regularization to encourage pruning by

\begin{equation}
\label{eq:reg}
    \mathcal{L}_{reg} =  \left( t - \frac{1}{N} \sum_{i=1}^{N} p(f_i) \right)^2
\end{equation}
where $t$ is a hyperparameter that represents a target activation rate~\cite{herrmann2020channel} at this layer. By adjusting $t$, we want to encourage the true activation to approach a fixed rate $t$.

The Softmax pruning method improves upon magnitude-based pruning in terms of inference latency because no sorting operation is needed. However, we empirically find the performance drop is more severe than that in the latter (See Table~\ref{tab:main_det}). We argue that the main cause of the drop lies in the discrepancy of pruned features. In training, we scale the features by a factor in a continuous range $p(f_i) \in (0,1)$, while during inference we enforce the factor to be drawn from a categorical binomial distribution. 

\paragraph{Gumbel Spatial Pruning.} To further bridge the gap between training and inference, 
a feasible solution is to sample a binary variable $z_i$ from the predicted Bernoulli distribution depicted by $p(f_i)$. Then, we send $z_i f_i$, instead of $p(f_i) f_i$ to the subsequent layers, making training and inference consistent. A straight-forward sampling could be $z_i = \arg\max \{p(f_i), 1-p(f_i)\}$. However, the $\arg\max$ operation is non-differentiable, preventing us from back-propagating gradients through $z_i$ to $p(f_i)$. 
Motivated by~\cite{jang2017categorical}, 
we propose to replace the Softmax operation with Gumbel Softmax, which provides a simple and efficient way to sample $z_i$.  
Specifically, we have
\begin{equation}
\label{eq:gumbel}
    \hat{p} ( f_i) = \frac{e^{\sigma_1 (f_i) + G_1} } {e^{\sigma_0 (f_i) + G_0} + e^{\sigma_1 (f_i)+ G_1} } 
\end{equation}
where $G_i=-\log(-\log(\epsilon_i))$ and $\epsilon_i$ are random noises drawn from uniform distribution $U(0,1)$. Note that Eq.~\ref{eq:gumbel} only ensures the soften sample $\hat{p} ( f_i)$ follows the Bernoulli distribution, while its value still holds non-discrete. In practice, we employ the hard estimation
\begin{equation}
    z_i = \mathds{1}  \left(\sigma_1(f_i) + G_1 > \sigma_0(f_i)+G_0\right) - \text{sg}(\hat{p}(f_i)) + \hat{p}(f_i)
\end{equation}
where $\mathds{1}(\cdot)$ is the indicator function, and $\text{sg}(\cdot)$ indicates stop gradient operation. During the forward pass $z_i$ is binary, and during back-propagation, the gradient is propagated straight-through since $\frac{\partial \mathcal{L}}{\partial z_i} = \frac{\partial \mathcal{L}}{\partial \hat{p} (f_i)}$. We term the proposed pruning layer as Gumbel Spatial Prunining (GSP) layers. Training of the GSP Layer also involves the sparse regularization in Eq.~\ref{eq:reg}, but replacing $p(f_i)$ with $z_i$. During inference, we index voxel features with $z_i=1$ instead of passing $z_i f_i$ to the next layer, avoiding unnecessary computation on zero vectors and therefore reducing costs.

\begin{table*}
    \centering
    \caption{\textbf{Results of TransL on nuScenes 3D detection validation set.} Compared to the TransL baseline, all pruning approaches remarkably reduce GFLOPs. SSP and GSP both achieve practical latency reduction while  MBP~\cite{chen2023voxelnext, liu2022spatial} fails. Among all methods, GSP shows the smallest perception accuracy drops in terms of NDS and mAP. Results are consistent for different numbers of sweeps.}
    \begin{tabular}{|c|c|c|c|c|c|cccccccccc|}
        \hline
         swps. & Methods &  GFLOPs & Latency & NDS & mAP & Car & Truck & Bus & Trail & Const & Ped & Moto & Bike & Cone & Bar\\
        \hline
         \multirow{4}*{10} &TransL &  128.7 & 305 & \textbf{67.7} & \textbf{62.8} & 85.6 & 60.1 & 72.1 & 37.4 & 26.4 & 85.0 & 69.9 & 54.0 & 72.8 & 65.8\\
         \cline{2-16}
         
         ~&TransL-MBP & \underline{71.6} & 336 & 66.5 & 60.5 & 85.9 & 58.1 & 70.7 & 35.5 & 22.5 & 85.5 & 65.5 & 42.9 & 71.2 & 70.6\\
         \cline{2-16}
         ~&TransL-SSP & 83.2 & \underline{268} & 66.8 & 60.5 & 85.8 & 48.4 & 71.1 & 39.2 & 25.0 & 84.7 & 63.9 & 48.2 & 70.7 & 66.0\\
         \cline{2-16}
         ~&TransL-GSP  & \textbf{57.4} & \textbf{225} & \underline{67.3} & \underline{62.0} & 85.6 & 57.8 & 68.9 & 37.6 & 25.2 & 84.4 & 66.0 & 51.5 & 71.2 & 71.7\\
        \hline
         \multirow{4}*{20} &TransL & 266.8 & 354 & \textbf{69.7} & \textbf{64.8} & 87.6 & 61.5 & 73.2 & 42.7 & 31.8 & 85.4 & 67.9 & 53.8 & 74.0 & 68.9\\
         \cline{2-16}
         ~&TransL-MBP  & \underline{124.5} & 403 & 67.2 & 61.2 & 85.0 & 60.3 & 71.4 & 39.6 & 24.3 & 85.7 & 66.9 & 46.7 & 72.9 & 71.2\\
         \cline{2-16}
         ~&TransL-SSP & 162.7 & \underline{291} & 68.0 & 61.5 & 87.0 & 50.1 & 72.0 & 39.9 & 26.7 & 85.3 & 65.1 & 49.2 & 72.0 & 67.3\\
         \cline{2-16}
         ~&TransL-GSP  & \textbf{124.3} & \textbf{250} & \underline{69.0} & \underline{64.5} & 86.7 & 60.2 & 72.0 & 38.9 & 25.8 & 86.1 & 69.9 & 58.8 & 74.5 & 71.8\\
        \hline
         \multirow{4}*{40} &TransL  & 615.9 & 445 & \textbf{70.9} & \textbf{65.5} & 88.3 & 63.0 & 73.5 & 43.3 & 33.5 & 86.3 & 70.1 & 54.2 & 75.9 & 69.0\\
         \cline{2-16}
         ~&TransL-MBP  & \underline{284.1} & 526 & 69.1 & 64.0 & 88.6 & 60.5 & 71.4 & 39.1 & 25.6 & 85.3 & 67.6 & 47.6 & 75.0 & 69.9\\
         \cline{2-16}
         ~&TransL-SSP  & 407.5 & \underline{379} & 69.6 & 64.7 & 87.9 & 61.5 & 73.5 & 42.1 & 31.6 & 85.9 & 67.4 & 53.1 & 74.5 & 68.3\\
         \cline{2-16}
         ~&TransL-GSP  & \textbf{281.0} & \textbf{292} & \underline{70.0} & \underline{65.2} & 88.8 & 59.2 & 73.6 & 39.5 & 27.9 & 88.4 & 70.7 & 56.2 & 75.8 & 72.0\\
    \hline
    \end{tabular}
    
    \label{tab:main_det}
\end{table*}

   


\section{Experiments}
\label{sec:experiments}


\subsection{Experimental Setups}

\noindent\textbf{Datasets and Metrics.} 
We perform an evaluation on the nuScenes~\cite{Caesar_2020_CVPR} dataset, focusing on two perception tasks: 3D object detection and BEV (Bird's-Eye-View) map segmentation. %
For assessing perception accuracy, we report the nuScenes Detection Score (NDS) and mean Average Precision (mAP) metrics for 3D object detection tasks, and mean Intersection over Union (mIoU) for BEV map segmentation tasks. 
Regarding efficiency improvements, we report the number of floating-point operations (FLOPs) and latency measured in milliseconds (ms) on an NVIDIA V100 GPU. Note FLOPs and latency are measured in the entire test set and we report the average number for a single forward pass.

\noindent \textbf{Baseline and Pruning Methods.} We list our baseline method and several pruning strategies that we compare with as follows.

\begin{itemize}
    \item \textbf{TransL:} The baseline architecture we used, adapted from the point cloud branch of TransFusion~\cite{transfusion}, which is composed of a VoxelNet~\cite{2017VoxelNet} encoder, a SECOND~\cite{second} decoder, and a transformer head.
    \item \textbf{TransL-MBP:} {M}agnitude {B}ased {P}runing method implemented in SPS-Conv~\cite{liu2022spatial} and VoxelNext~\cite{chen2023voxelnext}. We insert a pruning layer before each down-sampling layer of TransL as done in VoxelNeXt. The pruning rate is set to $[ 0.5,0.5,0.5,0] $ for the four pruning layers.
    \item \textbf{TransL-SSP:} Similar to TransL-MBP, except that the pruning layer is replaced with Softmax Spatial Pruning described in Section~\ref{sec:spatial_pruning}. Same pruning rate.
    \item \textbf{TransL-GSP:} Similar to TransL-SSP but the pruning layer is replaced with Gumbel Spatial Pruning.
\end{itemize}

\subsection{Main Results}

\noindent\textbf{3D Object Detection.} 
We compare the above methods in the validation set of the nuScenes 3D Object detection dataset.
Three configurations for the number of accumulated sweeps are considered, namely 10, 20, and 40. The corresponding results are listed in Tab.~\ref{tab:main_det}.

Several observations can be made from the results:

\begin{itemize}
    \item \textbf{TransL-MBP}: Compared to the baseline TransL, TransL-MBP achieves significant reductions in FLOPs by $44.4\%/53.4\%/53.9\%$ for $10/20/40$ sweeps, while maintaining a modest compromise in detection accuracy. Specifically, the NDS shows a decrease of $-1.2/-2.5/-1.8$, and mAP exhibits a decrease of $-2.3/-3.6/-1.5$. However, as discussed in Section~\ref{sec:spatial_pruning}, it is quite important to be aware that the FLOPs measurement does not account for the sorting operations. Consequently, when considering latency as a metric, there is no reduction but rather an increase. This phenomenon indicates that magnitude-based pruning does not yield acceleration benefits in practice.
    \item \textbf{TransL-SSP:} 
    In general, SSP outperforms MBP in both efficiency and perception accuracy. SSP does not rely on a sort operation so reductions in FLOPs and latency reach consistency. The latency is reduced by $12.1\%/17.8\%/14.8\%$ for $10/20/40$ sweeps. 
    \item \textbf{TransL-GSP:} The proposed GSP reaches a consistent reduction in GFLOPs and latency, yielding the best improvement with $26.2\%/29.4\%/34.4\%$ reduction in latency. Meanwhile, the performance accuracy drop is the smallest, with only $-0.4/-0.7/-0.9$ NDS and $-0.8/-0.3/-0.3$ mAP for $10/20/40$ sweeps.

\end{itemize}

We also record the results of TransL on the nuScenes test set, as shown in Tab. ~\ref{tab:test_det}. The results also show that the increase of the sweep number can improve the performance of the model, and the pruning method brings slight damage to the experimental metrics. 

\begin{table}[h]
    \centering
    \footnotesize
    \caption{\textbf{Results of TransL on nuScenes 3D detection test set.} }
    \begin{tabular}{|c|c|c|c|}
        \hline
         swps. & Methods & NDS & mAP\\
        \hline
         \multirow{2}*{10} & TransL & 69.3 & 64.4\\
         ~& TransL-GSP & 67.6 & 63.1\\
        \hline
         \multirow{2}*{20} & TransL & 70.6 & 66.0\\
         ~& TransL-GSP  & 69.2 & 64.6\\
        \hline
         \multirow{2}*{40} & TransL & 71.3 & 66.6\\
         ~& TransL-GSP & 70.2 & 65.5\\
        \hline    
    \end{tabular}
    \label{tab:test_det}
\end{table}

\begin{table*}[t]
    \centering
    \caption{\textbf{Results of TransL on nuScenes BEV map segmentation.} Similar conclusions to Tabel~\ref{tab:main_det} can be made.}
    \begin{tabular}{|c|c|c|c|c|cccccc|}
        \hline
          swps. &Methods & GFLOPs & Latency & mIoU & Drive & Ped & Walk & Stop & Park & Divide\\
        \hline
         \multirow{2}*{10} &TransL &  122.0 & 344 & 49.3 & 75.5 & 47.8 & 57.2 & 36.3 & 37.0 & 41.7\\
         \cline{2-11}
         ~ &TransL-GSP &  56.7 \sg{-53.5\%} & 273 \sg{-20.6\%} & 49.2 \sr{-0.1} & 75.6 & 47.3 & 56.5 & 35.9 & 36.3 & 41.6\\
        \hline
         \multirow{2}*{20} &TransL &  262.9 & 391 & 50.7 & 76.3 & 49.8 & 58.1 & 37.3 & 40.4 & 42.3\\
         \cline{2-11}
         ~ &TransL-GSP &  129.8 \sg{-50.6\%} & 298 \sg{-23.8\%} & 49.8 \sr{-0.9} & 76.0 & 49.2 & 57.4 & 36.9 & 37.7 & 41.8\\
        \hline
         \multirow{2}*{40} &TransL &  606.5 & 464 & 52.6 & 78.3 & 50.9 & 59.9 & 39.9 & 41.7 & 44.0\\
         \cline{2-11}
         ~ &TransL-GSP &  285.0 \sg{-53.0\%} & 335 \sg{-27.8\%} & 52.1 \sr{-0.5} & 77.3 & 51.3 & 59.5 & 39.2 & 41.9 & 43.5\\
         \hline
    
    \end{tabular}
    
    \label{tab:main_seg}
\end{table*}

In summary, the proposed TransL-GSP, utilizing a 40-sweep point cloud as input, achieves a faster runtime of $292$ ms per forward pass compared with the 10-sweep TransL baseline, which runs at $305$ ms. This improved efficiency is also accompanied by a notable enhancement in perception accuracy, as evidenced by the increase in NDS from $67.7$ to $70.0$ and mAP from $62.8$ to $65.2$. 
\textbf{Note that} here we report the latency of the whole network, in which a dense transformer head is also taken into account, while the latency reduction for the sole sparse encoder is much more obvious.


\noindent\textbf{BEV Map Segmentation.} 
In Table.~\ref{tab:main_seg}, we show a comparison between the TransL baseline and TransL-GSP on nuScenes BEV map segmentation task. 
The results are similar to the 3D detection task.
After implementing the proposed GSP layer, we reduce the latency of a 40-sweep model by $-27.8\%$,
achieving a running speed as fast as a 10-sweep TransL model. Meanwhile, the mIoU metric only drops by $-0.5\%$. 

\subsection{Ablation Studies}
\label{ablation}

\noindent \textbf{Generalization Ability.}
Besides the vanilla Transfusion method, we also implement the pruning strategy on LargeKernel3D~\cite{chen2023largekernel3d}, one of the SOTA point-cloud-based detection algorithms based on sparse convolution on the nuScenes dataset, in order to prove the generalization ability of our method. In Table.~\ref{tab:lk3d}, we show the comparison between the LK3D baseline and the LK3D-GSP on the nuScenes 3D detection task. The proposed GSP provides a consistent reduction in GFLOPs and latency, with a $28.4\%/23.5\%/30.3\%$ reduction in latency. Moreover, the performance accuracy drops are kept at a relatively low level. It proves the generalization ability of our strategy compared to other baseline methods. \textbf{Note that} the experiments on LargeKernel3D are implemented on an NVIDIA 3090ti GPU. 

\begin{table}[h]
    \centering
    \footnotesize
    \caption{\textbf{Results of LargeKernel3D on nuScenes 3D object detection test set.} }
    \begin{tabular}{|c|c|c|c|c|c|}
        \hline
         swps. & Methods &  GFLOPs & Latency & NDS & mAP\\
        \hline
         \multirow{2}*{10} & LK3D & 240 & 109 & 70.1 & 65.0\\
         \cline{2-6}
         ~&LK3D-GSP  & 179 \sg{-25.4\%} & 78 \sg{-28.4\%} & 68.1 & 63.5\\
        \hline
         \multirow{2}*{20} & LK3D & 451 & 132 & 71.3 & 65.9\\
         \cline{2-6}
         ~&LK3D-GSP  & 296 \sg{-34.4\%} & 101 \sg{-23.5\%} & 69.8 & 64.4\\
        \hline
         \multirow{2}*{40} & LK3D & 837 & 198 & 73.2 & 67.3\\
         \cline{2-6}
         ~&LK3D-GSP  & 516 \sg{-38.4\%} & 138 \sg{-30.3\%} & 71.6 & 66.0\\
        \hline
         
    \end{tabular}
 
    \label{tab:lk3d}
\end{table}



\begin{figure}
\centering
\includegraphics[width=\linewidth]{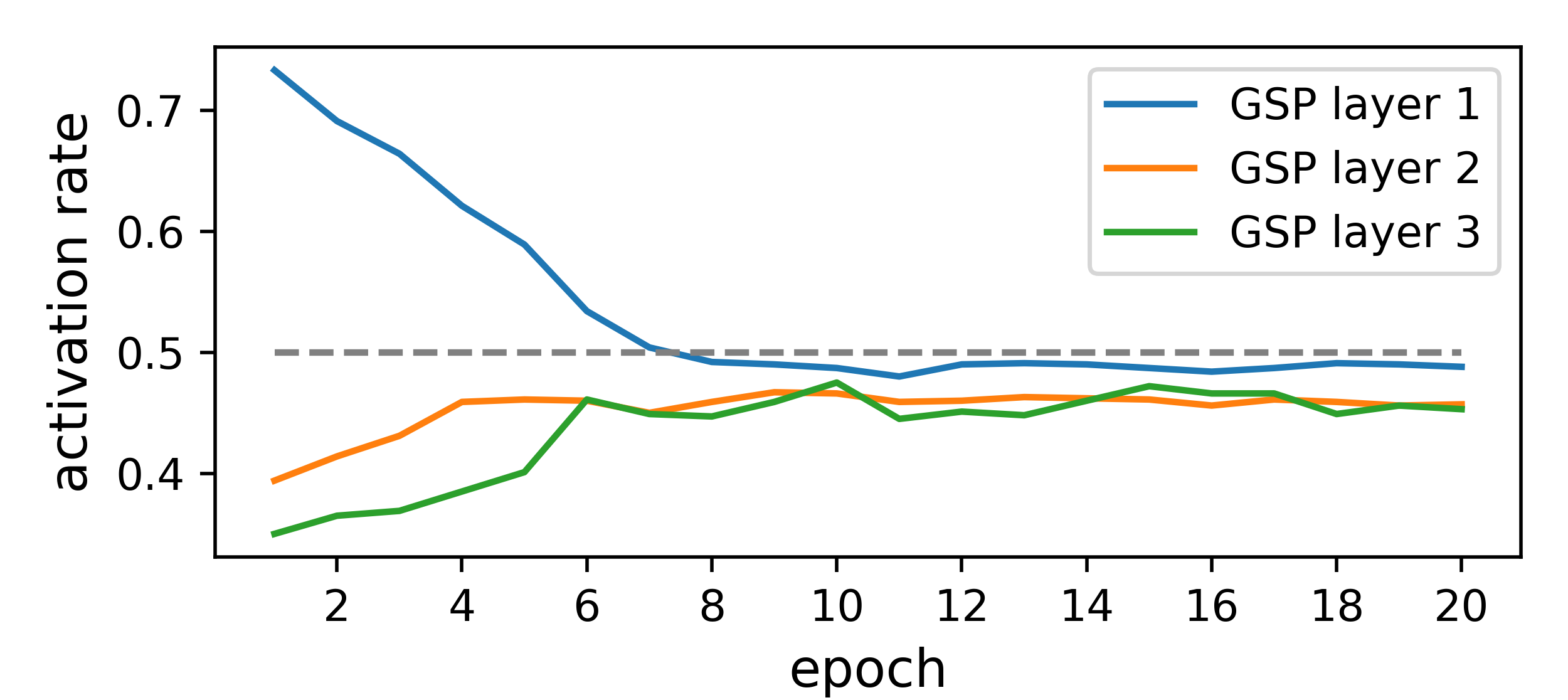}
\caption{\textbf{Dynamics of the actual activation rate.} Dashed line represents the target activation rate, $t=0.5$. }
\label{fig:epoch}
\end{figure}

\noindent\textbf{Dynamics of the Actual Activation Rate.} In Fig.~\ref{fig:epoch} we show how the actual activation rate evolves along with the training process. Two interesting findings can be found in the figure. Firstly, for all the GSP layers, the actual activation rate converges to the target $t$ at the 8th epoch. Secondly, we observe that for the first GSP layer, the activation rate keeps gradually reducing, while for the later two GSP layers, it increases. We speculate the reason may be that when the first layer prunes more and more, the latter two layers must try harder to find useful features so that the final perception performance does not degenerate, thus they have to activate more spatial features. 

\begin{table}[h]
    \centering
    \caption{Different target activation rate $t$ for nuScenes 3D object detection tasks}
    \begin{tabular}{|c|c|c|c|c|c|c|}
        \hline
        $t$ & 1 & 0.9 & 0.7 & 0.5 & 0.3 & 0.1\\
        \hline
        mAP & 62.8 & 62.6 & 62.4 & 62.0 & 59.6 & 48.7\\
        \hline
        NDS & 67.7 & 67.5 & 67.4 & 67.3 & 65.0 & 57.7\\
        \hline
        Latency & 305 & 272 & 241 & 225 & 206 & 182\\
        \hline
        GFLOPs & 128.7 & 99.9 & 75.6 & 57.4 & 42.1 & 16.8\\
        \hline
    \end{tabular}
\label{tab:ratio_det}
\end{table}

\begin{table}[h]
    \centering
    \caption{Different target activation rate $t$ for nuScenes BEV map segmentation tasks.}
    \begin{tabular}{|c|c|ccccc|}
        \hline
        $t$ & 1 & 0.9 & 0.7 & 0.5 & 0.3 & 0.1  \\
        \hline
        mIoU & 49.3 & 49.3 & 49.2 & 49.2 & 48.4 & 36.7\\
        \hline
        Latency & 344 & 337 & 301 & 273 & 249 & 219\\
        \hline
        GFLOPs & 122.0 & 109.2 & 83.6 & 56.7 & 29.7 & 6.9\\
    \hline
    \end{tabular}
    
    \label{tab:ratio_seg}
\end{table}

\begin{table*}[t]
    \centering
    \caption{\textbf{Latency breakdown.} We show how the computation cost is reduced for each layer of the sparse encoder. Throughout the downsampling process in each encoder layer, the reduction in latency becomes more salient. What is more, even considering the extra time consumed in the pruning process, the total latency is still reduced immensely. }
    \begin{tabular}{|c|c|c|c|c|c|c|c|c|}
        \hline
         \multirow{2}*{Sweep} &\multirow{2}*{Methods} &  \multirow{2}*{Conv Input} & \multicolumn{4}{c|}{Encoder Layers} & \multirow{2}*{Conv Output} & \multirow{2}*{Total}\\
         \cline{4-7}
         ~ & ~ & ~ & Layer 1 & Layer 2 & Layer 3 & Layer 4 & ~ & ~\\
        \hline
         \multirow{2}*{10} &TransL&  4.8 & 26.0 & 36.7 & 42.4 & 42.8 & 3.6 & 169.8\\
         \cline{2-9}
         ~ &TransL-GSP &  4.8 \sg{-0.0\%} & 27.8 \sr{+6.9\%} & 16.2 \sg{-55.9\%}& 15.7 \sg{-63.0\%} & 10.0 \sg{-76.6\%} & 2.8 \sg{-22.2\%} & 99.0 \sg{-41.7\%}\\
        \hline
         \multirow{2}*{20} &TransL&  6.8 & 36.9 & 49.5 & 52.0 & 49.3 & 3.8 & 204.3\\
         \cline{2-9}
         ~ &TransL-GSP &  6.8 \sg{-0.0\%} & 39.0 \sr{+5.7\%} & 18.3 \sg{-63.0\%} & 15.8 \sg{-69.6\%} & 10.6 \sg{-78.5\%} & 2.8 \sg{-26.3\%} & 112.9 \sg{-44.7\%}\\
        \hline
         \multirow{2}*{40} &TransL&  10.8 & 57.5 & 70.2 & 65.9 & 58.3 & 4.2 & 272.9\\
         \cline{2-9}
         ~ &TransL-GSP &  10.8 \sg{-0.0\%} & 59.3 \sr{+3.1\%} & 17.8 \sg{-74.6\%} & 15.3 \sg{-76.8\%} & 10.5 \sg{-82.0\%} & 2.7 \sg{-35.7\%} & 122.3 \sg{-58.8\%}\\
         \hline

    \end{tabular}
    
    \label{tab:latency_ana}
\end{table*}

\noindent\textbf{Trade-off between $t$ and Perception Performance.}
In general, a lower target activation rate $t$ results in a larger reduction in GFLOPs and latency, but potentially also a larger drop in perception performance. 
In Tab.~\ref{tab:ratio_det} and~\ref{tab:ratio_seg} we investigate how the performance varies with different activation rates $t$ for the nuScenes 3D object detection and BEV map segmentation tasks. 
For simplicity, we set an identical target $t$ for all the GSP layers.
We observe that with a $t>0.5$, the performance of the two perception tasks remains almost unchanged. When $t<0.5$, perception accuracy starts to degenerate, and when $t<0.1$ the drop becomes significant. 
In summary, $0.3<t<0.5$ is a good trade-off between efficiency and performance.

\noindent\textbf{Hard \textit{v.s.} Dynamic Voxelization.} 
In Table~\ref{tab:voxel}, we show the improvements of dynamic voxelization upon the original hard voxelization. We show the latency of both (1) the entire network and (2) the sparse encoder alone. 
With increasing point cloud size, the computation cost of hard voxelization becomes unaffordable, while that of dynamic voxelization only slightly increases.
Replacing hard voxelization with dynamic voxelization, the latency is reduced by $5.3\times$/$12.1\times$/$28.6\times$ for TransL backbone with 10/20/40 sweeps, and $9.5\times$/$22.7\times$/$56.7\times$ for TransL-GSP backbone with 10/20/40 sweeps. 

\begin{table}[h]
    \centering
    \caption{\textbf{Hard \textit{v.s.} Dynamic voxelization.}  
    The computation power consumption of the widely used hard voxelization grows prohibitively expensive when we use a large number of sweeps, \ie, 40. However, dynamic voxelization reduces the latency of the sparse backbone by $57\times$, showing its necessity for the use of multi-sweep point cloud.}
    \begin{tabular}{|c|c|c|c|c|}
    \hline
     \multirow{2}*{Swp.} &\multirow{2}*{Methods} &\multirow{2}*{Vox.}&  Network& Backbone\\
     & & & Latency  & Latency \\
    \hline
    \multirow{4}*{10}  &TransL & Hard &  1060 & 903\\
    \cline{2-5}
     ~ &TransL & Dyn. &  305 & 170\\
     \cline{2-5}
    ~ &TransL-GSP & Hard & 980 & 853\\
    \cline{2-5}
     ~ & TransL-GSP & Dyn. &\textbf{225} & \textbf{90}\\

     \hline
     \multirow{4}*{20} & TransL &Hard. & 2770 & 2637\\
     \cline{2-5}
     ~ &TransL&Dyn. &  354 & 218\\
     \cline{2-5}
     ~ &TransL-GSP&Hard&  2670 & 2563\\
     \cline{2-5}
     ~ &TransL-GSP&Dyn. &  \textbf{250} & \textbf{113}\\
     \hline
     \multirow{4}*{40} &TransL &Hard&  8710 & 8622\\
     \cline{2-5}
     ~ &TransL&Dyn. &  445 & 301\\
     \cline{2-5}
     ~ &TransL-GSP&Hard&  8610 & 8513\\
     \cline{2-5}
     ~ & TransL-GSP&Dyn. & \textbf{292} & \textbf{150}\\
    \hline
    \end{tabular}
    
    \label{tab:voxel}
\end{table}

\begin{figure}[h]
\centering
\includegraphics[width=\linewidth]{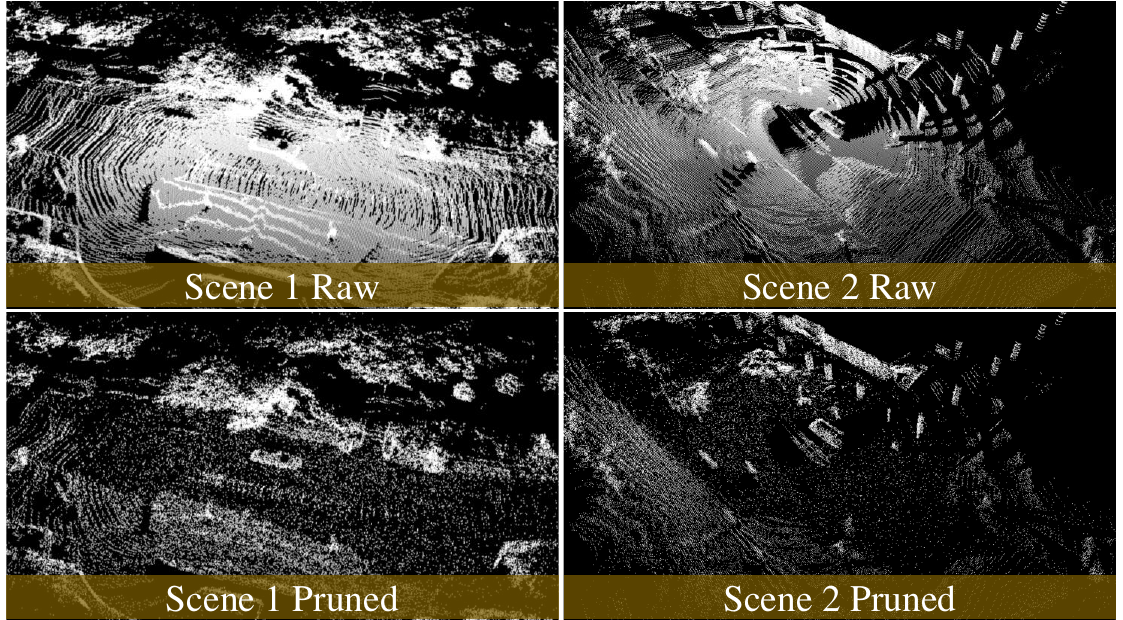}
\caption{\textbf{Visualization of GSP in point cloud view.} We visualize the raw point clouds and the GSP-pruned point clouds from two scenes. }
\label{fig:vis}
\end{figure}

\noindent\textbf{Visualizations.} 
As shown in Fig.~\ref{fig:vis}, to better understand the distribution of the points pruned by our strategy, we visualize point clouds before and after the pruning operation in two examples, scenes 1 and 2. We implement our pruning strategy because of the belief that it can improve the efficiency of the network. Besides, the strategy is expected to intuitively keep the useful information. As shown in the visualized examples, the target objects stand out from the background point clouds. This suggests that the remaining voxels seem to be more informative, whereas the pruned ones are less so. 

\noindent\textbf{Latency Breakdown.} 
In Table~\ref{tab:latency_ana}, we show a breakdown analysis of how the latency is reduced for each layer of the sparse encoder. 
Pruning starts from the end of Layer~1, and we observe that the latency reduction increases from Layer~2 to Layer~4, since every GSP layer prunes around $t= 50\%$ of its inputs. Consequently, the overall reduction of the entire sparse encoder exceeds $50\%$.

\begin{table}[h]
    \centering
    \caption{\textbf{Comparisons on different sampling techniques.} We implement different sampling methods on 3D detection and map segmentation tasks with the nuScenes dataset. }
    \begin{tabular}{|c|c|c|c|c|c|}
    \hline
         \multirow{2}*{Swp.} &\multirow{2}*{Methods} & \multirow{2}*{Strategy}& \multicolumn{2}{c|}{3D Detection} & Map Seg\\
         \cline{4-5}
         ~ & ~ &~ & NDS & mAP & mIoU\\
        \hline
         \multirow{4}*{10} &TransL & Even&  67.7 & 62.8 & 49.3\\
         \cline{2-6}
         ~ &TransL&Uneven &  68.0 & 62.9 & 49.6\\
         \cline{2-6}
         ~ &TransL-GSP &Even&  67.3 & 62.0 & 49.2\\
         \cline{2-6}
         ~ &TransL-GSP&Uneven &  67.8 & 62.3 & 50.2\\
        \hline
         \multirow{4}*{20} &TransL &Even&  69.7 & 64.8 & 50.7\\
         \cline{2-6}
        ~ & TransL &Uneven&  69.9 & 65.2 & 51.1\\
        \cline{2-6}
         ~ &TransL-GSP &Even&  69.0 & 64.5 & 49.8\\
         \cline{2-6}
         ~ &TransL-GSP &Uneven&  69.6 & 65.0 & 50.7\\
         \hline
 
    \end{tabular}
    
    \label{tab:sample}
\end{table}

\noindent\textbf{Sampling Strategy.} When faced with multiple sweeps of point clouds, we also tested our GSP on different sampling approaches. In nuScenes and Waymo, multiple sweeps are sampled evenly based on time stamps. However, we implement numerous experiments and figure out that uneven time distribution might contribute to tuning our models. We lower the sampling density of the sweeps with the furthest timestamps from that of the sample. Experimental results from Tab.~\ref{tab:sample} show that our sampling strategy achieves better qualitative results. The results indicate that under an appropriate sampling measure, the post-pruning performance even exceeds the pre-pruning performance, which proves the effectiveness of our GSP layer. 


\section{Conclusion}
\label{sec:conclusion}
In this paper, we observe the spatial sparsity in point cloud perception tasks and mitigate it by accumulating multiple point cloud sweeps, achieving promising performance in multiple tasks, including 3D object detection and BEV map segmentation. We then investigate the ascending time and computation power consumption and propose Gumbel Spatial Pruning layer to alleviate such issues. Extensive experiments prove our strategy to be both effective and efficient on various perception tasks, compared with the vanilla backbone. We hope that our work will stimulate more related research in the future. 








{\small
\bibliographystyle{IEEEtran}
\bibliography{root}

@inproceedings{2017VoxelNet,
	title={VoxelNet: End-to-End Learning for Point Cloud Based 3D Object Detection},
	author={ Zhou, Yin  and  Tuzel, Oncel },
	booktitle={2018 IEEE/CVF Conference on Computer Vision and Pattern Recognition (CVPR)},
	year={2018},
	pages= {4490-4499}
}

@inproceedings{2019PointRCNN,
	title={PointRCNN: 3D Object Proposal Generation and Detection From Point Cloud},
	author={ Shi, Shaoshuai  and  Wang, Xiaogang  and  Li, Hongsheng },
	booktitle={2019 IEEE/CVF Conference on Computer Vision and Pattern Recognition (CVPR)},
	year={2019},
	pages= {770-779}
}

@InProceedings{2020Point,
	author = {Shi, Weijing and Rajkumar, Ragunathan (Raj)},
	title = {Point-GNN: Graph Neural Network for 3D Object Detection in a Point Cloud},
	booktitle = {The IEEE Conference on Computer Vision and Pattern Recognition (CVPR)},
	month = {June},
	pages= {1711-1719},
	year = {2020}
}

@INPROCEEDINGS {10204261,
	author = {N. Zhang and Z. Pan and T. H. Li and W. Gao and G. Li},
	booktitle = {2023 IEEE/CVF Conference on Computer Vision and Pattern Recognition (CVPR)},
	title = {Improving Graph Representation for Point Cloud Segmentation via Attentive Filtering},
	year = {2023},
	pages = {1244-1254},
	month = {June}
}

@InProceedings{lu2023link,
	author    = {Lu, Tao and Ding, Xiang and Liu, Haisong and Wu, Gangshan and Wang, Limin},
	title     = {LinK: Linear Kernel for LiDAR-Based 3D Perception},
	booktitle = {Proceedings of the IEEE/CVF Conference on Computer Vision and Pattern Recognition (CVPR)},
	month     = {June},
	year      = {2023},
	pages     = {1105-1115}
}

@article{2020Sparse,
	title={Sparse Single Sweep LiDAR Point Cloud Segmentation via Learning Contextual Shape Priors from Scene Completion},
	author={ Yan, Xu  and  Gao, Jiantao  and  Li, Jie  and  Zhang, Ruimao  and  Li, Zhen  and  Huang, Rui  and  Cui, Shuguang },
	year={2020},
}

@ARTICLE{2021barrera,
	author={Barrera, Alejandro and Beltrán, Jorge and Guindel, Carlos and Iglesias, José Antonio and García, Fernando},
	journal={IEEE Access}, 
	title={BirdNet+: Two-Stage 3D Object Detection in LiDAR Through a Sparsity-Invariant Bird’s Eye View}, 
	year={2021},
	volume={9},
	number={},
	pages={160299-160316},
	doi={10.1109/ACCESS.2021.3131389}}

@InProceedings{Liu_2023_ICCV,
	author    = {Liu, Yibo and Zhu, Kelly and Wu, Guile and Ren, Yuan and Liu, Bingbing and Liu, Yang and Shan, Jinjun},
	title     = {MV-DeepSDF: Implicit Modeling with Multi-Sweep Point Clouds for 3D Vehicle Reconstruction in Autonomous Driving},
	booktitle = {Proceedings of the IEEE/CVF International Conference on Computer Vision (ICCV)},
	month     = {October},
	year      = {2023},
	pages     = {8306-8316}
}

@INPROCEEDINGS{wang_2022_cvpr,
	author={Wang, Jun and Li, Xiaolong and Sullivan, Alan and Abbott, Lynn and Chen, Siheng},
	booktitle={2022 IEEE/CVF Conference on Computer Vision and Pattern Recognition Workshops (CVPRW)}, 
	title={PointMotionNet: Point-Wise Motion Learning for Large-Scale LiDAR Point Clouds Sequences}, 
	year={2022},
	volume={},
	number={},
	pages={4418-4427},
	doi={10.1109/CVPRW56347.2022.00488}}

@InProceedings{Li_2022_CVPR,
	author    = {Li, Yingwei and Yu, Adams Wei and Meng, Tianjian and Caine, Ben and Ngiam, Jiquan and Peng, Daiyi and Shen, Junyang and Lu, Yifeng and Zhou, Denny and Le, Quoc V. and Yuille, Alan and Tan, Mingxing},
	title     = {DeepFusion: Lidar-Camera Deep Fusion for Multi-Modal 3D Object Detection},
	booktitle = {Proceedings of the IEEE/CVF Conference on Computer Vision and Pattern Recognition (CVPR)},
	month     = {June},
	year      = {2022},
	pages     = {17182-17191}
}

@misc{chen2023largekernel3d,
	title={LargeKernel3D: Scaling up Kernels in 3D Sparse CNNs}, 
	author={Yukang Chen and Jianhui Liu and Xiangyu Zhang and Xiaojuan Qi and Jiaya Jia},
	year={2023},
	eprint={2206.10555},
	archivePrefix={arXiv},
	primaryClass={cs.CV}
}

@inproceedings{liu2022bevfusion,
	title={BEVFusion: Multi-Task Multi-Sensor Fusion with Unified Bird's-Eye View Representation},
	author={Liu, Zhijian and Tang, Haotian and Amini, Alexander and Yang, Xingyu and Mao, Huizi and Rus, Daniela and Han, Song},
	booktitle={IEEE International Conference on Robotics and Automation (ICRA)},
	year={2023}
}

@InProceedings{Jiao_2023_CVPR,
	author    = {Jiao, Yang and Jie, Zequn and Chen, Shaoxiang and Chen, Jingjing and Ma, Lin and Jiang, Yu-Gang},
	title     = {MSMDFusion: Fusing LiDAR and Camera at Multiple Scales With Multi-Depth Seeds for 3D Object Detection},
	booktitle = {Proceedings of the IEEE/CVF Conference on Computer Vision and Pattern Recognition (CVPR)},
	month     = {June},
	year      = {2023},
	pages     = {21643-21652}
}

@InProceedings{Caesar_2020_CVPR,
	author = {Caesar, Holger and Bankiti, Varun and Lang, Alex H. and Vora, Sourabh and Liong, Venice Erin and Xu, Qiang and Krishnan, Anush and Pan, Yu and Baldan, Giancarlo and Beijbom, Oscar},
	title = {nuScenes: A Multimodal Dataset for Autonomous Driving},
	booktitle = {Proceedings of the IEEE/CVF Conference on Computer Vision and Pattern Recognition (CVPR)},
	month = {June},
	year = {2020}
}

@InProceedings{Sun_2020_CVPR,
	author = {Sun, Pei and Kretzschmar, Henrik and Dotiwalla, Xerxes and Chouard, Aurelien and Patnaik, Vijaysai and Tsui, Paul and Guo, James and Zhou, Yin and Chai, Yuning and Caine, Benjamin and Vasudevan, Vijay and Han, Wei and Ngiam, Jiquan and Zhao, Hang and Timofeev, Aleksei and Ettinger, Scott and Krivokon, Maxim and Gao, Amy and Joshi, Aditya and Zhang, Yu and Shlens, Jonathon and Chen, Zhifeng and Anguelov, Dragomir},
	title = {Scalability in Perception for Autonomous Driving: Waymo Open Dataset},
	booktitle = {Proceedings of the IEEE/CVF Conference on Computer Vision and Pattern Recognition (CVPR)},
	month = {June},
	year = {2020}
}

@InProceedings{Yi_2023_ICCV,
	author    = {Yi, Xuanyu and Deng, Jiajun and Sun, Qianru and Hua, Xian-Sheng and Lim, Joo-Hwee and Zhang, Hanwang},
	title     = {Invariant Training 2D-3D Joint Hard Samples for Few-Shot Point Cloud Recognition},
	booktitle = {Proceedings of the IEEE/CVF International Conference on Computer Vision (ICCV)},
	month     = {October},
	year      = {2023},
	pages     = {14463-14474}
}

@article{yi2024mvgamba,
  title={MVGamba: Unify 3D Content Generation as State Space Sequence Modeling},
  author={Yi, Xuanyu and Wu, Zike and Shen, Qiuhong and Xu, Qingshan and Zhou, Pan and Lim, Joo-Hwee and Yan, Shuicheng and Wang, Xinchao and Zhang, Hanwang},
  journal={arXiv preprint arXiv:2406.06367},
  year={2024}
}

@article{liang2024pointmamba,
  title={Pointmamba: A simple state space model for point cloud analysis},
  author={Liang, Dingkang and Zhou, Xin and Xu, Wei and Zhu, Xingkui and Zou, Zhikang and Ye, Xiaoqing and Tan, Xiao and Bai, Xiang},
  journal={arXiv preprint arXiv:2402.10739},
  year={2024}
}

@inproceedings{liu2022spatial,
	title={Spatial Pruned Sparse Convolution for Efficient 3D Object Detection},
	author={Liu, Jianhui and Chen, Yukang and Ye, Xiaoqing and Tian, Zhuotao and Tan, Xiao and Qi, Xiaojuan},
	booktitle={Advances in Neural Information Processing Systems},
	year={2022}
}

@inproceedings{centerpoint,
	title={Center-based 3d object detection and tracking},
	author={Yin, Tianwei and Zhou, Xingyi and Krahenbuhl, Philipp},
	booktitle={Proceedings of the IEEE/CVF conference on computer vision and pattern recognition},
	pages={11784--11793},
	year={2021}
}

@inproceedings{transfusion,
	title={Transfusion: Robust lidar-camera fusion for 3d object detection with transformers},
	author={Bai, Xuyang and Hu, Zeyu and Zhu, Xinge and Huang, Qingqiu and Chen, Yilun and Fu, Hongbo and Tai, Chiew-Lan},
	booktitle={Proceedings of the IEEE/CVF conference on computer vision and pattern recognition},
	pages={1090--1099},
	year={2022}
}

@inproceedings{huang2023cp3,
	title={CP3: Channel Pruning Plug-In for Point-Based Networks},
	author={Huang, Yaomin and Liu, Ning and Che, Zhengping and Xu, Zhiyuan and Shen, Chaomin and Peng, Yaxin and Zhang, Guixu and Liu, Xinmei and Feng, Feifei and Tang, Jian},
	booktitle={Proceedings of the IEEE/CVF Conference on Computer Vision and Pattern Recognition},
	pages={5302--5312},
	year={2023}
}

@article{second,
	title={Second: Sparsely embedded convolutional detection},
	author={Yan, Yan and Mao, Yuxing and Li, Bo},
	journal={Sensors},
	volume={18},
	number={10},
	pages={3337},
	year={2018},
	publisher={MDPI}
}

@inproceedings{chen2023voxelnext,
	title={Voxelnext: Fully sparse voxelnet for 3d object detection and tracking},
	author={Chen, Yukang and Liu, Jianhui and Zhang, Xiangyu and Qi, Xiaojuan and Jia, Jiaya},
	booktitle={Proceedings of the IEEE/CVF Conference on Computer Vision and Pattern Recognition},
	pages={21674--21683},
	year={2023}
}

@misc{zhou2019endtoend,
	title={End-to-End Multi-View Fusion for 3D Object Detection in LiDAR Point Clouds}, 
	author={Yin Zhou and Pei Sun and Yu Zhang and Dragomir Anguelov and Jiyang Gao and Tom Ouyang and James Guo and Jiquan Ngiam and Vijay Vasudevan},
	year={2019},
	eprint={1910.06528},
	archivePrefix={arXiv},
	primaryClass={cs.CV}
}

@misc{qi2023ocbev,
	title={OCBEV: Object-Centric BEV Transformer for Multi-View 3D Object Detection}, 
	author={Zhangyang Qi and Jiaqi Wang and Xiaoyang Wu and Hengshuang Zhao},
	year={2023},
	eprint={2306.01738},
	archivePrefix={arXiv},
	primaryClass={cs.CV}
}

@misc{wang2023uni3detr,
	title={Uni3DETR: Unified 3D Detection Transformer}, 
	author={Zhenyu Wang and Yali Li and Xi Chen and Hengshuang Zhao and Shengjin Wang},
	year={2023},
	eprint={2310.05699},
	archivePrefix={arXiv},
	primaryClass={cs.CV}
}

@misc{jang2017categorical,
	title={Categorical Reparameterization with Gumbel-Softmax}, 
	author={Eric Jang and Shixiang Gu and Ben Poole},
	year={2017},
	eprint={1611.01144},
	archivePrefix={arXiv},
	primaryClass={stat.ML}
}

@InProceedings{Zhou_2021_CVPR,
	author    = {Zhou, Zixiang and Zhang, Yang and Foroosh, Hassan},
	title     = {Panoptic-PolarNet: Proposal-Free LiDAR Point Cloud Panoptic Segmentation},
	booktitle = {Proceedings of the IEEE/CVF Conference on Computer Vision and Pattern Recognition (CVPR)},
	month     = {June},
	year      = {2021},
	pages     = {13194-13203}
}

@inproceedings{jiang2021guided,
	title={Guided point contrastive learning for semi-supervised point cloud semantic segmentation},
	author={Jiang, Li and Shi, Shaoshuai and Tian, Zhuotao and Lai, Xin and Liu, Shu and Fu, Chi-Wing and Jia, Jiaya},
	booktitle={Proceedings of the IEEE/CVF international conference on computer vision},
	pages={6423--6432},
	year={2021}
}

@inproceedings{ge2023metabev,
	title={Metabev: Solving sensor failures for 3d detection and map segmentation},
	author={Ge, Chongjian and Chen, Junsong and Xie, Enze and Wang, Zhongdao and Hong, Lanqing and Lu, Huchuan and Li, Zhenguo and Luo, Ping},
	booktitle={Proceedings of the IEEE/CVF International Conference on Computer Vision},
	pages={8721--8731},
	year={2023}
}

@inproceedings{liu2022prototype,
	title={Prototype-voxel contrastive learning for LiDAR point cloud panoptic segmentation},
	author={Liu, Minzhe and Zhou, Qiang and Zhao, Hengshuang and Li, Jianing and Du, Yuan and Keutzer, Kurt and Du, Li and Zhang, Shanghang},
	booktitle={2022 International Conference on Robotics and Automation (ICRA)},
	pages={9243--9250},
	year={2022},
	organization={IEEE}
}

@inproceedings{guan2022m3detr,
	title={M3detr: Multi-representation, multi-scale, mutual-relation 3d object detection with transformers},
	author={Guan, Tianrui and Wang, Jun and Lan, Shiyi and Chandra, Rohan and Wu, Zuxuan and Davis, Larry and Manocha, Dinesh},
	booktitle={Proceedings of the IEEE/CVF winter conference on applications of computer vision},
	pages={772--782},
	year={2022}
}

@article{fan2022fully,
	title={Fully sparse 3d object detection},
	author={Fan, Lue and Wang, Feng and Wang, Naiyan and ZHANG, ZHAO-XIANG},
	journal={Advances in Neural Information Processing Systems},
	volume={35},
	pages={351--363},
	year={2022}
}

@article{ye2022lidarmutlinet,
	title={LidarMutliNet: Unifying LiDAR Semantic Segmentation, 3D Object Detection, and Panoptic Segmentation in a Single Multi-task Network},
	author={Ye, Dongqiangzi and Chen, Weijia and Zhou, Zixiang and Xie, Yufei and Wang, Yu and Wang, Panqu and Foroosh, Hassan},
	journal={arXiv preprint arXiv:2206.11428},
	year={2022}
}

@article{li2023frame,
	title={Frame Fusion with Vehicle Motion Prediction for 3D Object Detection},
	author={Li, Xirui and Wang, Feng and Wang, Naiyan and Ma, Chao},
	journal={arXiv preprint arXiv:2306.10699},
	year={2023}
}

@inproceedings{herrmann2020channel,
	title={Channel selection using gumbel softmax},
	author={Herrmann, Charles and Bowen, Richard Strong and Zabih, Ramin},
	booktitle={European Conference on Computer Vision},
	pages={241--257},
	year={2020},
	organization={Springer}
}

@article{tang2022torchsparse,
	title={Torchsparse: Efficient point cloud inference engine},
	author={Tang, Haotian and Liu, Zhijian and Li, Xiuyu and Lin, Yujun and Han, Song},
	journal={Proceedings of Machine Learning and Systems},
	volume={4},
	pages={302--315},
	year={2022}
}

@article{sun2023trosd,
  title={Trosd: A new rgb-d dataset for transparent and reflective object segmentation in practice},
  author={Sun, Tianyu and Zhang, Guodong and Yang, Wenming and Xue, Jing-Hao and Wang, Guijin},
  journal={IEEE Transactions on Circuits and Systems for Video Technology},
  volume={33},
  number={10},
  pages={5721--5733},
  year={2023},
  publisher={IEEE}
}

@inproceedings{li2025segment,
  title={Segment, Lift and Fit: Automatic 3D Shape Labeling from 2D Prompts},
  author={Li, Jianhao and Sun, Tianyu and Wang, Zhongdao and Xie, Enze and Feng, Bailan and Zhang, Hongbo and Yuan, Ze and Xu, Ke and Liu, Jiaheng and Luo, Ping},
  booktitle={European Conference on Computer Vision},
  pages={407--423},
  year={2025},
  organization={Springer}
}

@article{sun2024diffusion,
  title={Diffusion-Based Depth Inpainting for Transparent and Reflective Objects},
  author={Sun, Tianyu and Hu, Dingchang and Dai, Yixiang and Wang, Guijin},
  journal={IEEE Transactions on Circuits and Systems for Video Technology},
  year={2024},
  publisher={IEEE}
}

@article{xie2023part,
  title={Part-guided 3d rl for sim2real articulated object manipulation},
  author={Xie, Pengwei and Chen, Rui and Chen, Siang and Qin, Yuzhe and Xiang, Fanbo and Sun, Tianyu and Xu, Jing and Wang, Guijin and Su, Hao},
  journal={IEEE Robotics and Automation Letters},
  volume={8},
  number={11},
  pages={7178--7185},
  year={2023},
  publisher={IEEE}
}

@article{hu2025variation,
  title={Variation-Robust Few-Shot 3D Affordance Segmentation for Robotic Manipulation},
  author={Hu, Dingchang and Sun, Tianyu and Xie, Pengwei and Chen, Siang and Yang, Huazhong and Wangy, Guijin},
  journal={IEEE Robotics and Automation Letters},
  year={2025},
  publisher={IEEE}
}

@misc{jin2025angledomainguidancelatent,
      title={Angle Domain Guidance: Latent Diffusion Requires Rotation Rather Than Extrapolation}, 
      author={Cheng Jin and Zhenyu Xiao and Chutao Liu and Yuantao Gu},
      year={2025},
      eprint={2506.11039},
      archivePrefix={arXiv},
      primaryClass={cs.LG},
      url={https://arxiv.org/abs/2506.11039}, 
}

@article{sun2026mvanimate,
  title={MVAnimate: Enhancing Character Animation with Multi-View Optimization},
  author={Sun, Tianyu and Fu, Zhoujie and Zhang, Bang and Lin, Guosheng},
  journal={arXiv preprint arXiv:2602.08753},
  year={2026}
}
}

\end{document}